\documentclass[letterpaper, 10 pt, conference]{ieeeconf}  % Comment this line out if you need a4paper
\usepackage{cite}
\usepackage{amsmath,amssymb,amsfonts}
\usepackage{graphicx}
\usepackage{threeparttable}
\usepackage{booktabs}
\usepackage{makecell} % allow \\ in cells
\usepackage{multirow} % allow one cell with multiple rows
\usepackage{textcomp}
\usepackage[colorlinks,linkcolor=blue,citecolor=black]{hyperref}
\usepackage{algpseudocode}
\usepackage{algorithm}
\usepackage{xcolor}

\IEEEoverridecommandlockouts                              % This command is only needed if 
\title{\LARGE \bf
Preparation of Papers for IEEE Sponsored Conferences \& Symposia*
}

\title{\LARGE \bf Deep Domain Adaptation Regression for \\ Force Calibration of Optical Tactile Sensors  }
\author{Zhuo Chen$^{1}$, Ni Ou$^{1,2}$, Jiaqi Jiang$^{1}$ and Shan Luo$^{1}$
\thanks{*This work was supported by the EPSRC project ``ViTac: Visual-Tactile Synergy for Handling Flexible Materials" (EP/T033517/2).}
\thanks{$^{1}$Zhuo Chen, Ni Ou, Jiaqi Jiang and Shan Luo are with the Robot Perception Lab, Centre for Robotics Research, Department of Engineering, King's College London, London WC2R 2LS, United Kingdom. Emails: {\tt\small \{zhuo.7.chen, shan.luo \}@kcl.ac.uk}.}
\thanks{$^{2}$Ni Ou is with the State Key Laboratory of Intelligent Control and Decision of Complex Systems, Beijing Institute of Technology, Beijing, 100081, China.}
}
\begin{document}
\maketitle
\thispagestyle{empty}
\pagestyle{empty}

%%%%%%%%%%%%%%%%%%%%%%%%%%%%%%%%%%%%%%%%%%%%%%%%%%%%%%%%%%%%%%%%%%%%%%%%%%%%%%%%
\begin{abstract}
Optical tactile sensors provide robots with rich force information for robot grasping in unstructured environments. The fast and accurate calibration of three-dimensional contact forces holds significance for new sensors and existing tactile sensors which may have incurred damage or aging. However, the conventional neural-network-based force calibration method necessitates a large volume of force-labeled tactile images to minimize force prediction errors, with the need for accurate Force/Torque measurement tools as well as a time-consuming data collection process. To address this challenge, we propose a novel deep domain-adaptation force calibration method, designed to transfer the force prediction ability from a calibrated optical tactile sensor to uncalibrated ones with various combinations of domain gaps, including marker presence, illumination condition, and elastomer modulus. Experimental results show the effectiveness of the proposed unsupervised force calibration method, with lowest force prediction errors of 0.102N (3.4\% in full force range) for normal force, and 0.095N (6.3\%) and 0.062N (4.1\%) for shear forces along the x-axis and y-axis, respectively. This study presents a promising, general force calibration methodology for optical tactile sensors. 
\end{abstract}

%%%%%%%%%%%%%%%%%%%%%%%%%%%%%%%%%%%%%%%%%%%%%%%%%%%%%%%%%%%%%%%%%%%%%%%%%%%%%%%%
\section{INTRODUCTION}\label{intro}
Optical tactile sensors \cite{GelSight1} with high sensitivity in perceiving object geometry, slip and position have now been widely studied. With the superiority of directly capturing tactile information with high-resolution images, optical tactile sensors like GelSight~\cite{GelSight0}, GelTip~\cite{geltip} and Digit~\cite{digit} significantly enhance the capabilities of robots, particularly when integrated with machine learning models. Among those touch sensations applicable for measurement by optical tactile sensors, force sensing stands out as it is crucial for monitoring dynamic contact status and providing feedback for robot control~\cite{rc1} and in-hand manipulation~\cite{rc2}.

% However, force calibration in optical tactile sensing faces two primary challenges: firstly, the dependency on extensive labeled data for model training, and secondly, the sensor components may undergo alterations that could affect the force predictions. For example, the soft elastomers are key components to detect contact forces and the hyperelastic nature of soft elastomers results in a highly non-linear relationship between the light gradient changes in tactile images and the applied forces~\cite{trans_tatile}, and their degragation will result into changes in the force predictions. 

% To estimate force values from tactile images, deep neural networks are often employed instead of physical models to achieve precise force predictions, necessitating a substantial volume of tactile images paired with labeled forces~\cite{insight}. This exhaustive data collection process have to be reiterated for new sensors or existing ones undergo changes in physical properties due to damages or aging. As a result, the deployment of force calibration models is hindered, especially for users of commercial tactile sensors where accurate calibration tools like the Nano17 F/T sensors are unavailable.

Force calibration in optical tactile sensing faces three primary challenges. Firstly, deep neural networks are often preferred over physical models to estimate forces from tactile images, necessitating a substantial volume of tactile images paired with labeled forces for model training~\cite{insight}. This demanding data collection process has to be repeated for new sensors. As a result, deploying force calibration models, especially for users of commercial tactile sensors lacking accurate calibration tools like the Nano17 F/T sensors, becomes challenging. Secondly, alternations in sensor components, such as degradations of soft elastomers, can lead to inaccuracies in force predictions. Thirdly, differences in marker presence/distributions, illumination conditions and elastomer modulus in different optical tactile sensors prevent force prediction models from being adapted from calibrated sensors to uncalibrated ones. 

	\begin{figure}[!t]
            \centering
            \includegraphics[width=0.95\linewidth]{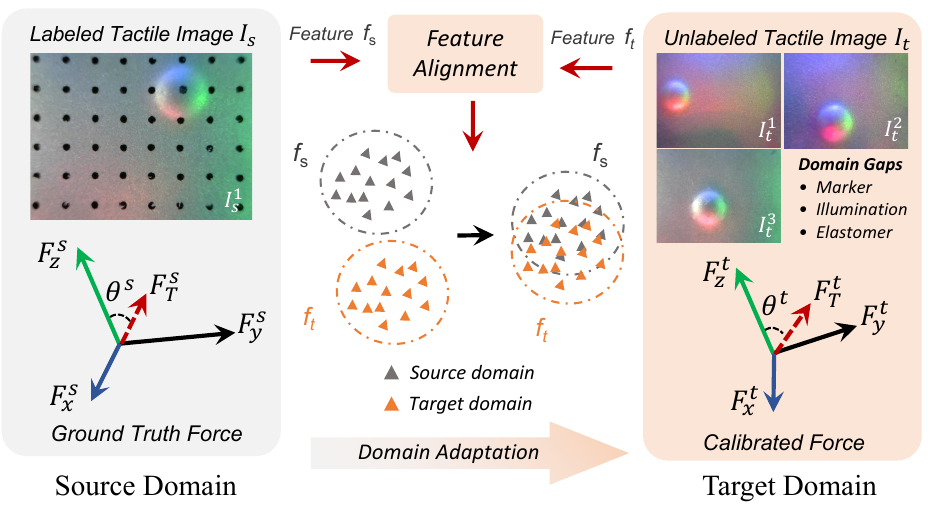}
		% \caption{Deep domain adaptation regression for force calibration of optical tactile sensors. Upon completing the calibration process, the unlabeled tactile images $I_t$ from target domain are aligned with labeled tactile images $I_s$ from source domain, with ground truth force $F^s$, in the feature space. Meanwhile, the unlabeled sensors ($I_t^1,I_t^2,I_t^3$), each exhibiting different combinations of domain gaps  can predict calibrated forces $F^t$. Here, $F_T$ represents the total force, $F_z$ denotes the normal force, $F_x$ and $F_y$ denotes the shear forces, $\theta$ represents the force angle.}
  \caption{Deep domain adaptation for force calibration of optical tactile sensors. Upon completing domain adaptation, the feature space of the unlabeled tactile images $I_t$ from the target domain and the labeled tactile images $I_s$ from the source domain are aligned. A shared regressor trained with ground truth forces $F^s$ in source domain can be used to predict $F^t$ in target domain. $I_t^1$, $I_t^2$ and $I_t^3$ denote tactile images in target domain with varying combinations of domain gaps. $F_T$ represents the total force, $F_z$ denotes the normal force, $F_x$ and $F_y$ denote the shear forces, and $\theta$ represents the force angle, respectively.}
		\label{Fig1}
	\end{figure}
% On the other hand, three domain gaps—marker presence, illumination condition, and elastomer modulus—exist among new sensors or old sensors, significantly deteriorate the generalization performance of force prediction models adapated from the trained sensors in source domain to another sensors in the target domain. Specifically, the presence of marker equips maker-based optical sensors with notable advantages in slip detection and force visualization, whereas it also disrupts the continuity of tactile images collected from sensors without marker. Furthermore, illumination condition, particularly regarding RGB color and light intensity, tends to vary due to fluctuations in physical parameters such as resistances, LED installation angles, and membrane reflectivity. Even minor changes in illumination conditions can yield distinct alterations in RGB channels of tactile images. Moreover, achieving consistent mechanical properties in gel elastomers, such as Young's modulus, proves challenging, even within the same fabrication batch. In regular usage, elastomer surfaces are also susceptible to wear, tear, and aging, exacerbating the degradation of force-prediction performance in neural network models, necessitating re-calibration. 

Transfer learning presents a promising approach for models trained on existing sensors (source domain) to adapt to new sensors (target domain). However, current transfer-learning methods used for this task, such as fine-tuning~\cite{trans_tatile}, are supervised, which requires extensive labeled force information from both domains. Hence, there remains a significant demand for a dedicated model utilizing unsupervised transfer learning methods in the force calibration of optical tactile sensors.

In this study, we propose a novel domain adaptation regression model to address the unsupervised force calibration challenge in optical tactile sensors, illustrated in Fig.~\ref{Fig1}. This approach can eliminate the need for costly force/torque measurement tools in force calibration of optical tactile sensors and significantly reduce the calibration time through domain adaptation. Experimental results demonstrate the successful adaptation of pretrained models on force-labeled tactile images to sensors with diverse domain gaps and unlabeled images. To our best knowledge, this is the first work that utilizes the deep domain adaptation regression method to address this challenge. %Additionally, this method holds promise for adaptation to other regression problems in robot tactile sensing, such as pose estimation and contact depth estimation of in-hand objects. 
The contributions of this work are summarized as follows:
% \begin{enumerate}
%     \item We introduce a deep domain adaptation regression method to address the challenge of unsupervised force calibration in optical tactile sensors. 
%     \item We investigate the impact of different combinations of three key domain gaps on domain adaptation performance and verify the effectiveness of our model.
%     \item We release an open-source GelSight image \href{https://github.com/Zhuochenn/DAR_OTS/releases/download/dataset/data.zip}{dataset} annotated with forces across multiple domains, which could serve as both source domain (for training) and target domain (for evaluation) in this task. 
% \end{enumerate}
\begin{enumerate}
    \item A novel domain adaptation regression method is introduced to address the challenge of unsupervised force calibration in optical tactile sensors;
    \item The impact of different combinations of three key domain gaps on domain adaptation performance is investigated, and the effectiveness of our model is verified;
    \item \href{https://github.com/Zhuochenn/DAR_OTS/releases/download/dataset/data.zip}{A dataset} has been made publicly accessible for unsupervised force calibration of optical tactile sensors. 
\end{enumerate}

The rest of the paper is structured as follows: Section~\ref{sec:relatedworks} provides an overview of related works; Section~\ref{sec:methodology} introduces our methodology; Section~\ref{sec:data} details our data collection and implementation; Section~\ref{sec:experiment} analyses the experimental results. Finally, Section~\ref{sec:conclusion} presents the discussion and summarises the work.

\begin{figure*}[t]
\centering
\includegraphics[width=\textwidth]{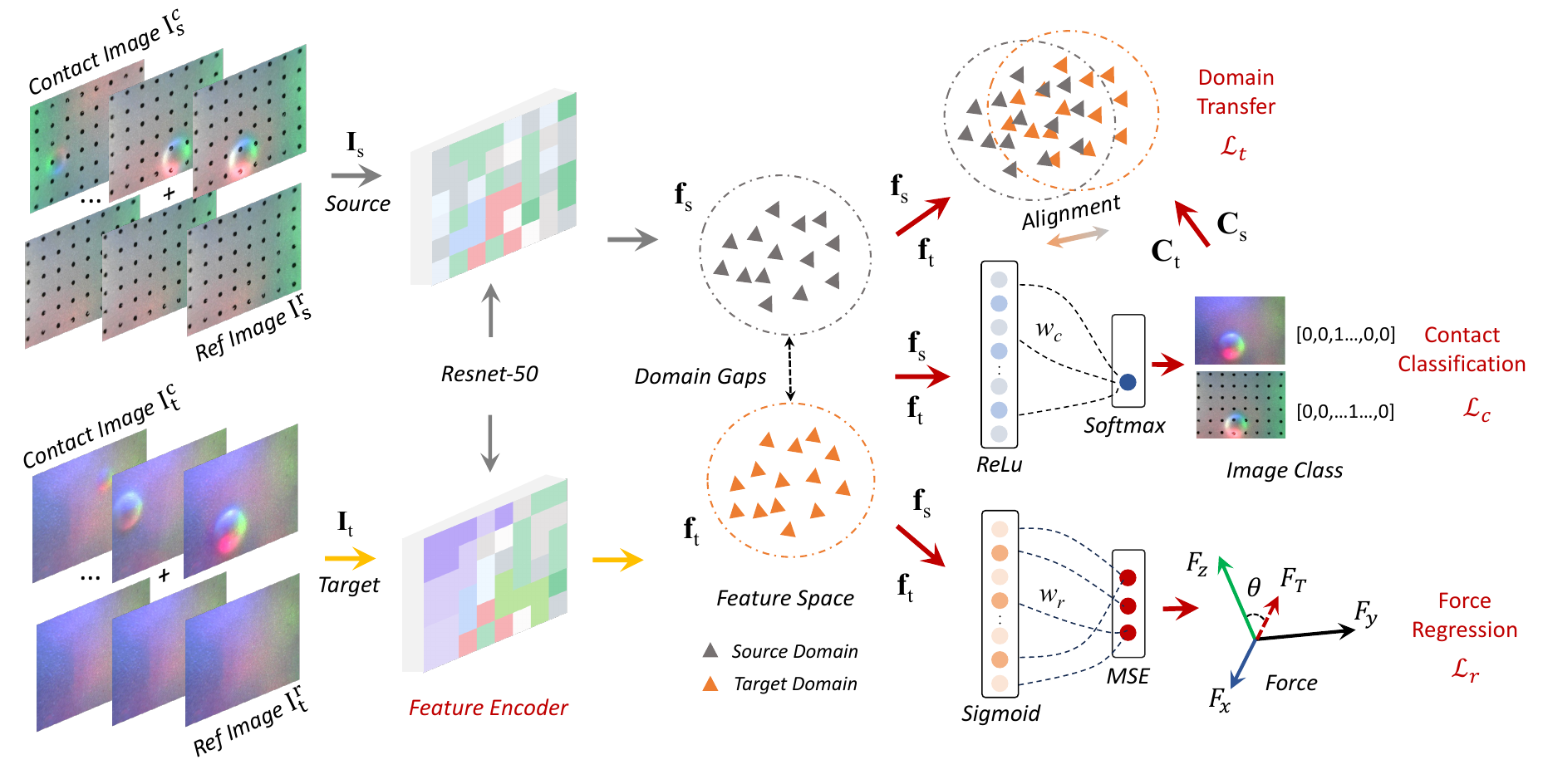}
\caption{Deep domain adaptation regression model for force calibration. The model comprises four components: a feature encoder, a contact classification head, a force regression head, and a domain transfer head. The model takes batches of tactile images from the source domain $\mathbf{I}_s$ and the target domain $\mathbf{I}_t$ as inputs. Both $\mathbf{I}_s$ and $\mathbf{I}_t$ consist of concatenated tactile images comprising a current contact image $\textup{I}^c$ and a reference image $\textup{I}^r$. The domain transfer head accepts features ($\mathbf{f}_s$ and $\mathbf{f}_t$) and ground truth contact labels $\mathbf{C}_s$ from the source domain, as well as the pseudo labels $\mathbf{C}_t$ predicted from the classifier. The overall loss $\mathcal{L}_h$ is weighted sum of the regression loss $\mathcal{L}_r$, classification loss $\mathcal{L}_c$ and domain adaptation loss $\mathcal{L}_t$.}
\label{fig2}
\end{figure*}
\section{RELATED WORK}
\label{sec:relatedworks}
\subsection{Optical Tactile Sensors}\label{R1}

Optical tactile sensors are characterized by utilizing compact cameras to capture high-resolution images of deformed soft elastomers when contacting with objects. By exploiting image processing techniques or machine learning models, tactile images can be used for various downstream robotic manipulation tasks, including force estimation~\cite{insight}, slip detection~\cite{slip-detect}, object recognition~\cite{texturerecog}, and object localization~\cite{localization}. Optical tactile sensors can be  broadly categorized into marker-motion-based sensors like TacTip~\cite{geltip} and high-resolution image-based sensors like GeSight~\cite{GelSight1}. TacTip-like sensors consist of physical pins on the sensing elastomer and primarily detect shear information and contact position by capturing marker displacement. In contrast, GelSight-like sensors utilize RGB cameras to capture images with detailed textures and reconstruct contact height maps using photometric stereo methods. The black marker dots can also be painted on the gel elastomer for shear displacement estimation~\cite{uvtac}. Over the past decades, GelSight-like sensors have been developed into different variants,  such as GelSlim~\cite{gelslim3}, GelWedge~\cite{gelwedge} and GelSvelte~\cite{gelsvelte}. In this work, we primarily focus on force calibration using GelSight sensors, but it's worth noting that this method is also applicable to TacTip-like sensors and other types of optical tactile sensors.

\subsection{Force Calibration of Optical Tactile Sensors}\label{R2}

Force calibration of optical tactile sensors involves establishing the mapping relationship between tactile images and contact forces~\cite{GelSight1}. This process can be achieved either by applying a physical-based model/mechanical calibration method to compute contact force, or by directly constructing an end-to-end image-to-force mapping using a deep neural network~\cite{insight}. However, due to the hyperelastic properties of gel elastomer membranes, the mechanical calibration method, which assumes linear elastic behavior, is only accurate for small deformations, and its force prediction performance significantly deteriorates with large deformations~\cite{hyperelastic}. In contrast, data-driven deep neural networks exhibit impressive performance in non-linear regression tasks and can maintain accurate force prediction across the full working range of deformation~\cite{insight}. Nevertheless, the neural network-based approach is data-driven and sensitive to the feature distribution of tactile images, necessitating data collection and model re-training whenever the sensor's physical factors change. There is a pressing need for a method that simplify the cumbersome data collection process and expedites the calibration process on new or aging sensors by adapting trained models on existing sensors .

\subsection{Domain Adaptation Regression}\label{R3}
Domain adaptation~\cite{DA1} aims at mitigating the distribution shifts between the labeled source domain and the unlabeled target domain. Adversarial domain adaptation techniques, such as DANN\cite{DANN} and Deep-CORAL\cite{Coral}, are widely used to learn an embedding feature space where the source and target domains cannot be distinguished. While instance-based methods primarily correct the shift by re-weighting source instances or minimizing specific distances between the two distributions, such as KL-divergence\cite{KL} and Maximum Mean Discrepancy (MMD)\cite{MMD}. However, these methods are typically applied to classification\cite{DSAN} and segmentation \cite{seg} tasks and are not directly applicable to regression problems. While recent methods like RSD\cite{RSD} and DARE-GRAM\cite{GRAM} have been proposed to tackle domain adaptation regression problems, they have only been tested on a few benchmark datasets with simple domain gaps. In contrast, our domain adaptation regression task for force calibration will consider different combinations of domain gaps in optical tactile sensors, which is more complex and problem-specific.

\section{METHODOLOGY}
\label{sec:methodology}
\subsection{Problem Definition}\label{m.1}

In the domain adaptation regression problem for the force calibration of optical tactile sensors, we are provided with a labeled tactile image dataset $\mathcal{D}_s =\{(\mathbf{I}^i_s,(\mathbf{F}^i_s,\mathbf{C}^i_s))\}^{n_s}_{i=1}$ including the source domain with $n_s$ samples, and an unlabeled dataset $\mathcal{D}_t =\{(\mathbf{I}^i_t) \}^{n_t}_{i=1}$ with $n_t$ samples. Here, $\mathbf{I}^i_s$ and $\mathbf{I}^i_t$ represent the tactile images from the source domain and the target domain, respectively. $\mathbf{F}^i_s = (F^i_{x},F^i_{y},F^i_{z})_s$ denotes the ground truth of the applied normal force vector $F^i_{z}$ and shear forces vector $F^i_{x}$ and $F^i_{y}$ in source domain, while $\mathbf{C}^i_s$ indicates the contact class of the tactile images in the source domain, as described in Section~\ref{dataCollect}. 

The primary challenge in this task is that $\mathcal{D}_s$ and $\mathcal{D}_t$ are sampled from different sensors or the same sensor with varying marker presence, illumination conditions, or elastomer modulus. Consequently, tactile images collected from sensors with different combinations of domain gaps exhibit distinct feature distributions, i.e., $P(\mathbf{I}_s) \neq P(\mathbf{I}_t)$, whereas the objective is to learn a shared regressor $h:\mathbf{I} \rightarrow \mathbf{F}$ capable of directly mapping the $i_{\textup{th}}$ unlabeled tactile images $\mathbf{I}^i_t$ to calibrated forces vector $\mathbf{F}^i_t$ in the target domain. Therefore, our goal is to minimize the mean force prediction error in the target domain:

\begin{equation}\label{Eq.1}
\underset{h}\arg \min\mathbb{E}_{(\mathbf{I}^i_t,\mathbf{F}^i_t)} \left \lVert h(\mathbf{I}^i_t),\mathbf{F}^i_t \right \lVert_2^2
\end{equation}

However, since we only possess labeled images in the source domain, this optimization problem described in Equation \ref{Eq.1} is reformulated to minimize the Mean Square Error (MSE) between the predicted force and ground truth forces on the labeled source samples. Additionally, we aim to minimize the domain distribution discrepancy between the source domain and target domain:

\begin{equation}\label{Eq.2}
     \underset{h}\arg \min\lambda_r\underbrace{\frac{1}{n_s}\underset{i=1}{\overset{n_s}{\sum}}\left \Vert \hat{\mathbf{F}^i_s}-\mathbf{F}^i_s\right\Vert_2^2}_{\mathcal{L}_r} +\lambda_t\underbrace{\hat{d}_\mathcal{H}(\mathbf{f}_s,\mathbf{C}_s,\mathbf{f}_t,\mathbf{C}_t)}_{\mathcal{L}_t}
\end{equation}

where $\lambda_r\geq0$, $\lambda_t\geq0$ represent the weights assigned to the regression loss $\mathcal{L}_r$ and domain transfer loss $\mathcal{L}_t$. $\hat{\mathbf{F}^i_s}$ denotes the predicted forces using source images, while $\mathbf{f}_s$ and $\mathbf{f}_t$ denote the features extracted from the source domain and target domain, respectively. $\hat{d}_\mathcal{H}(\cdot,\cdot)$ is the esitimation of Local Maximum Mean Discrepancy (LMMD) \cite{DSAN}, with $\mathcal{H}$ being the Reproducing Kernel Hillbert Space (RKHS). This equation aims to learn a shared feature space and a shared regressor simultaneously so that forces in images from both the source or target domains can be predicted.

To compute the domain transfer loss $\mathcal{L}_t$, we refer to the Deep Subdomain Adaptation Network (DSAN)\cite{DSAN}, which has demonstrated the capability to capture fine-grained information for categories and align relevant domain distributions to learn a shared classifier. Hence, we try to reduce the discrepancy between source domain and target domain by optimizing the following function:

\begin{equation}\label{Eq.3}
\underset{g}\arg\min\lambda_c\underbrace{\frac{1}{n_s}\overset{n_s}{\underset{i=1}{\sum}} J(g(\mathbf{I}^i_s),\mathbf{C}^i_s)}_{\mathcal{L}_c}+\lambda_t\hat{d}_\mathcal{H}(\mathbf{f}_s,\mathbf{C}_s,\mathbf{f}_t,\mathbf{C}_t)
\end{equation}
where $\lambda_c\geq0$ represents the weight of classification loss $\mathcal{L}_c$, $g(\cdot)$ is a classifier for tactile images with different contact class, and $J(\cdot,\cdot)$ denotes the cross-entropy error. Compared to other domain adaptation regression methods that use a single regression head, the classification head included in this equation not only provides essential pseudo contact class labels $\mathbf{C}_t$ for the unlabeled images to calculate the LMMD but also benefits the regression task, as shown in \cite{clfhelpreg}.

Then, by combining Equations \ref{Eq.2} and \ref{Eq.3}, we obtain the overall loss function $\mathcal{L}_h$ of our model:

\begin{equation}\label{Eq.4}
    \mathcal{L}_h = \lambda_r\mathcal{L}_r + \lambda_c\mathcal{L}_c + \lambda_t\mathcal{L}_t
\end{equation}

By optimizing the loss function in Equation \ref{Eq.4}, a shared regressor can be learned for both the source domain and target domain while maintaining a well-aligned feature space. 

\subsection{Deep Domain Adaptation Regression Network}
\label{m.2}

Based on Section \ref{m.1}, the deep domain adaptation regression network in this task can be divided into four components as depicted in Fig.~\ref{fig2}: (1) feature encoder, (2) contact classification head, (3) force regression head and (4) domain transfer head. Firstly, the inputs from each domain consist of the concatenation of tactile images with contact $\textup{I}^c$ and reference tactile images without contact $\textup{I}^r$, which are then fed into the feature encoder. The feature encoder with a ResNet-50 backbone and a linear bottleneck layer jointly extracts the deep representations $\mathbf{f}_s$ and $\mathbf{f}_t$ of tactile images from the source domain $\mathbf{I}_s$ and the target domain $\mathbf{I}_t$, respectively. Next, the contact classification head composed of a single linear layer with a ReLU activation function classifies the contact cases of tactile images in the target domain and provides pseudo labels $\mathbf{C}_t$ for target images, which are then used for LMMD calculation with the ground truth labels $\mathbf{C}_s$ of source images. The force regression head using a linear layer with a Sigmoid activation function minimizes the MSE between the predicted forces $\mathbf{\hat{F}}_s$ and the ground truth forces $\mathbf{F}_s$. The domain transfer head aligns two domains in the feature space by minimizing LMMD. 

By using this model, tactile images $\mathbf{I}_s$, contact forces $\mathbf{F}_s$ and contact class $\mathbf{C}_s$ from existing sensors can be leveraged as source domain, while adapting existing models to new sensors or aging sensors by just collecting unlabeled tactile images $\mathbf{I}_t$ as target domain. This process could eliminate the need for force measurement tools, such as F/T sensors, during the calibration stage. Furthermore, the new model can be trained with fewer epochs by adapting existing models, thereby reducing the time required compared to the traditional supervised force calibration process.

\section{DATA COLLECTION AND IMPLEMENTATION}\label{DCID}
\label{sec:data}

 As reported in Section \ref{intro}, marker presence, illumination condition, and gel elastomer are three essential domain variables of different GelSight sensors, which are denoted as $wm/m$, $i$ and $b$ respectively, as shown Table~\ref{Domain_Gap_Define}. For data collection, three different elastomers and illumination conditions indexed by 0, 1, 2 are used. The elastomer $b$ indexed with higher number is with higher hardness. To study the domain adaptation performance between sensors with different combinations of these variables, we obtain four types of labeled tactile images, i.e., $mb_0i_0$, $wmb_0i_0$, $wmb_1i_1$ and $wmb_2i_2$, and pair them into nine domain adaptation groups listed in Table~\ref{Domain_Gap_Define}. For example, in the case of $mb_0i_0\rightarrow wmb_0i_0$, tactile images $\mathbf{I_s}$ are with markers and collected with elastomer-0 and illumination-0, while $\mathbf{I}_t$ are without markers and collected with elastomer-1 and illumination-1. It is worth noting that, for data collection, we only need to collect $mb_0i_0$, $wmb_1i_1$ and $wmb_2i_2$ in real GelSight sensors with corresponding physical properties. While $wmb_0i_0$ is generated by applying an inpainting method \cite{TELEA} on the $mb_0i_0$, which ensures only the marker-presence gap exists between the two domains. 

\begin{figure}[t]
\centering
\includegraphics[width=0.45\textwidth]{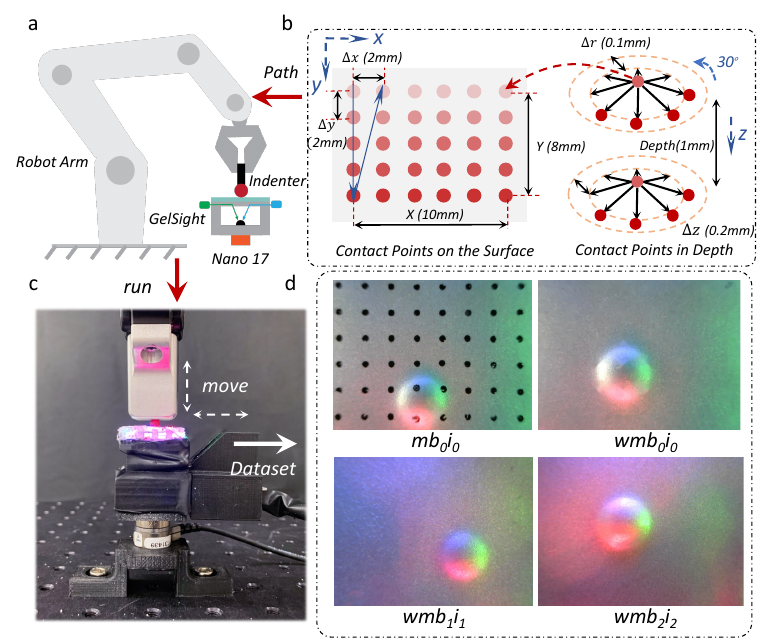}
\caption{(a) Data collection setup with a robot arm, a GelSight sensor, a sphere indenter and a Nano17 F/T sensor. (b) Programmed contact path for data collection. Contact pixel denotes a point for indentation. (c) Real-world setup. (d) Four groups of collected tactile images with different combinations of domain gaps.}
\label{fig3}
\end{figure}

\begin{table}[htbp]
\caption{Domain Adaptation Group}
 \centering
 \begin{tabular}{c|c}
      \toprule
      \#Variable & Domain Adaptation Group\\
      \midrule
       one & $mb_0i_0 \rightarrow wmb_0i_0$, $wmb_0i_0 \rightarrow mb_0i_0$\\
       two & \makecell{$wmb_0i_0 \rightarrow wmb_1i_1$, $wmb_0i_0 \rightarrow wmb_2i_2$ \\ $wmb_1i_1 \rightarrow wmb_0i_0$, $wmb_1i_1 \rightarrow wmb_2i_2$}\\
       three & \makecell{$mb_0i_0 \rightarrow wmb_1i_1$, $mb_0i_0 \rightarrow wmb_2i_2$ \\ $wmb_1i_1 \rightarrow mb_0i_0$}\\%
      \bottomrule
     \end{tabular}
     \begin{tablenotes}
         \item[1] * $m/wm$: with/without markers
         \item[2] * $b_0, b_1, b_2$: elastomer - 0, 1, 2
         \item[3] * $i_0, i_1, i_2$: illumination condition - 0, 1, 2
     \end{tablenotes}
     \label{Domain_Gap_Define}
\end{table}

\subsection{Data Collection}\label{dataCollect}

As shown in Fig.~\ref{fig3}a, the data collection setup comprises a UR5e robotic arm with a two-finger Robotiq gripper, a sphere indenter (d=3 mm), a flat-surface GelSight sensor, and a Nano17 F/T sensor. The robot arm is programmed through MoveIt to follow a planned path, as depicted in Fig.~\ref{fig3}b. The camera inside the GelSight sensor is synchronized with the Nano17 sensor to capture tactile images and contact forces simultaneously. 

Regarding the indenter's contact, 6$\times$5 surface points are pre-determinated in GelSight's 10$\times$8 mm$^2$ surface. For each contact point, the indenter is pressed into different depths ranging from 0 to 1 mm with a step of 0.2 mm. To collect shear forces, the motion of the robotic arm is divided into two stages: moving downwards and then moving horizontally shown in Fig.~\ref{fig3}c. When moving at a specific depth, the indenter follows the motion rules in Algorithm~\ref{motion}.  Specifically, it moves in a cyclical radial motion around a fixed radius towards 12 different angles. After the 12 angles of motions finished, the radius of the circle increases with a step of 0.1 mm from 0.1 mm to 0.6 mm, and the next cycle begins. There are $k=30$ contact points on the surface, and $n=361$ contact points in depth (including one reference position before contact). All the tactile images are collected from a force range of -3 N to 0 N for normal force in the $z$-axis and a range of -0.75 N to 0.75 N for shear forces in the $x$-axis and $y$-axis. This is limited by the maximum thickness of gel elastomer (3 mm) and the non-slip displacement ($<$0.6 mm) between the sphere indenter and the elastomer surface. As described in Section \ref{m.1}, we introduce contact class $\mathbf{C}^i_s$ in the loss function. Here, we allocate the contact class for each tactile image based on the index of the contact points in depth, i.e., $\mathbf{C}^i_s\in \mathbb{R}^{361}$ is a one-hot vector representing the label of $\mathbf{I}^i_s$, where $C^{ij}_s=1$ denotes $\mathbf{I}^i_s$ labeled with class $j$. The contact class is only used in training stage, which will not hinder the prediction of continuous force values in test.

\begin{algorithm}[htbp]
\caption{Indenter's Contact Motion}\label{motion}
\begin{algorithmic}[1]
\State \textbf{Input}: Number of surface contact points $k$, surface contact points $P=\{P_1,...,P_k\}$, where $P_i=\{p_{1},...,p_{n}\}$ contains all contact points in depth, $p_{i}=\{x,y,z\}$ denotes the coordinate of a contact point.
\Procedure{Move To Contact Points}{$P,k$}
    \For{$i=0...k$} 
        \State $p_0 \leftarrow P_i[0]$ \Comment{ origin point $p_0$  without contact}
        \State $moveTo(p_0)$ 
        \For{$j=1...n$}
            \State $p_1 \leftarrow (p_0[0],p_0[1],P_{ij}[2]), p_2 \leftarrow P_{ij}$
            \State $moveTo(p_1)$ \Comment{Move downwards}
            \State $moveTo(p_2)$ \Comment{Move horizontally}
            \State $imageCapture(), forceRecord()$
            \State $moveTo(p_0)$ \Comment{Move to origin}
        \EndFor
    \EndFor
\EndProcedure
\end{algorithmic}
\end{algorithm}

Fig.~\ref{fig3}d illustrates the collected tactile images within four groups. It is evident that $mb_0i_0$, $wmb_1i_1$, and $wmb_2i_2$ exhibit distinctive illumination conditions, also corresponding to three real GelSight sensors with different elastomer modulus. The data group $wmb_0i_0$ is directly generated from $mb_0i_0$ via the inpainting method, which controls the variables of illumination and elastomer, and shares the same ground truth force with $mb_0i_0$. Notably, we have three groups ($mb_0i_0$, $wmb_0i_0$, $wmb_1i_1$) that comprise a total of $3 \times 10,830$ images collected by the path in Fig.~\ref{fig3}b, along with $873$ images of $wmb_2i_2$. The last group of $wmb_2i_2$, containing only 873 images, is collected along a sparser path to test the performance of our model when trained with a few unlabeled images as target domain, while the other three data groups with ten times more images are utilized to evaluate the model performance across different domains with the same data size. These four data groups hold promise for being leveraged as source domains to extend our domain adaptation model to new sensors by simply collecting unlabeled tactile images without using force measurement tools.

\subsection{Implementation \& Evaluation Metrics}\label{iem}

In the training stage, three models are first trained with data groups $mb_0i_0$, $wmb_0i_0$, and $wmb_1i_1$, respectively, where only the regression layer is used for source-domain supervision ($\lambda_r=1, \lambda_c=0, \lambda_t=0$), and an initial learning rate $\eta_0 = 0.1$ for 20 epochs. This step aims at offering existing models for domain adaptation. In real-world use, we usually have an existing sensor with the trained model, and then the model and tactile images are used as source domain in this task for adapting to new/old sensors in target domain. For the second training stage, the corresponding pretrained model of source domain is selected and adapted by adding transfer loss and classification loss, i.e., $\lambda_r=1, \lambda_c=1, \lambda_t=1$ with an initial learning rate $\eta_0 = 0.01$ for 10 epochs. The learning rate in the backbone layer is always set as 10 times smaller than the initial learning rate. SGD optimizer is used with a momentum of 0.9, along with a learning rate scheduler $\eta = \eta_0 \cdot (1+0.0003 \cdot i)^{-0.75}$, where $i$ is the number of iterations. The batch size is set as 32, and all experiments are trained on a NVIDIA RTX 3090 GPU. 

All source-domain data are used in training stage, while the target-domain tactile images are split into train, valid and test dataset with a ratio of 0.6:0.2:0.2. The ground truth forces $\mathbf{F}_s$ from the source domain are normalized using min-max normalization to the range of $[0,1]$ in the training stage. For testing, the model exclusively receives unlabeled tactile image batches from the target domain. The calibrated force $\mathbf{F}_t$ is then derived from the regression head and subsequently subject to de-normalization for error calculation. The evaluation metrics used in Section \ref{experiment} are the Mean Absolute Error (MAE) and the coefficient of determination $R^2$, while tSNE is used to visualize domain distances in the feature space.

\section{EXPERIMENT RESULTS \& ANALYSIS}\label{experiment}
\label{sec:experiment}
In this section, we demonstrate the domain adaptation performance of our model in force calibration of GelSight sensors across different domain adaptation groups as shown in Table \ref{Domain_Gap_Define}. The baseline of this task is the source-only method, which directly predicts forces in the target domain using the pretrained models from source domain. Two other representative domain adaptation methods, including DANN\cite{DANN} and GRAM\cite{GRAM}, are also compared with our method in subsections \ref{one gap}, \ref{two gaps} and \ref{three gaps}. The force prediction errors in Tables \ref{TableOneGap}, \ref{TableTwoGaps}, and \ref{Table4} are calculated by the average of MAE with shear forces and normal forces.

\subsection{Marker Presence}\label{one gap}

\begin{figure*}[htbp]
\centering
\includegraphics[width=1\textwidth]{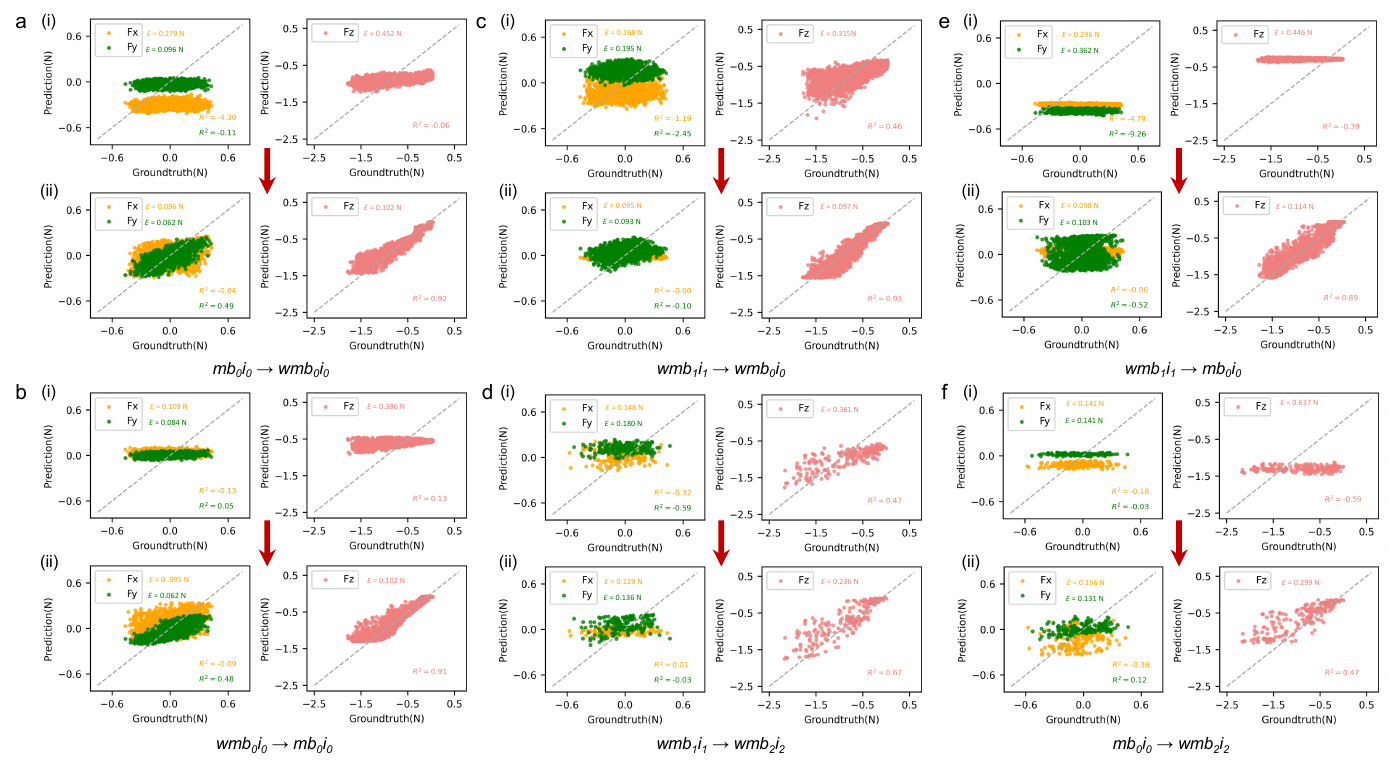}
\caption{Force prediction errors compared among domain adaptation groups featuring one variable (a-b), two variables (c-d), and three variables (e-f). This comparison is conducted using both the source-only method (i) and our domain adaptation method (ii).}
\label{force_errors}
\end{figure*}

The GelSight sensors with markers offer significant advantages in slip detection and force visualization, while the sensors without markers show advantages in object classification. Given the wide use of both types of sensors, it is desired to directly transfer the trained force prediction model from one of those to its counterpart. However, if we employ the source-only method directly from $mb_0i_0$ to $wmb_0i_0$, as depicted in Fig.~\ref{force_errors}a-i and Fig.~\ref{force_errors}b-i, the force prediction performance is unacceptable, with $R^2$ values of -4.3, -0.11, and -0.06 in the $xyz$-axis, respectively. The error $E$ for normal force can even reach up to 0.452N (15\% of the force range in the $z$-axis), while for the shear forces, the errors are 0.279 N (18.6\%) and 0.096 N (6.4\%), respectively. Conversely, in the transition from $wmb_0i_0$ to $mb_0i_0$, the $R^2$ values are -0.13, 0.05, and 0.13 in the $xyz$-axis, and the MAE errors are high. These results confirm the existence of significant gaps of marker presence, leading to substantial errors of force prediction across different domains.

After applying our method, as depicted in Fig.~\ref{force_errors}a-ii and Fig.~\ref{force_errors}b-ii, the errors $E$ of normal force in both groups decrease to $0.102$ N (3.4\%), while the $R^2$ values increase to 0.92 and 0.91 respectively. This represents an increase in accuracy of more than 10\% after domain adaptation. Although the $R^2$ values for shear force still appear low, the MAE errors in the $x$-axis decrease by 12.2\% from 0.279 N (18.6\%) to 0.096 N (6.4\%) in the $mb_0i_0 \rightarrow wmb_0i_0$ transition, which is acceptable when compared with the shear force error of around 0.025 N in the supervised model. One possible reason for the low $R^2$ values in the shear force could be that the collected shear forces are too small, mostly lying in the range of -0.75 N to 0.75 N, compared with the normal forces in -3 N to 0 N. Therefore, the static tactile images with small deformations contain negligible information for the shear forces, resulting in poorer transfer performance than normal forces. Regarding the comparison results with DANN and GRAM, our method shows an advantage in the overall average error (0.086 N), as shown in Table \ref{TableOneGap}.

\begin{table}[htbp]
    \caption{Average Force Prediction Error with \textbf{One} variable\\(Marker, Unit $N$)\\}
    \centering
    \begin{threeparttable}
        \begin{tabular}{c|cc|c}
            \toprule
            Method & \makecell{$mb_0i_0$ \\ $\rightarrow$ \\ $wmb_0i_0$}& \makecell{$wmb_0i_0$ \\ $\rightarrow$ \\ $mb_0i_0$}& Avg \\
            \midrule
            source-only & 0.276 & 0.196 &0.236\\
            \midrule
            DANN & 0.098 & 0.118 &0.108\\
                GRAM &\textbf{0.082}  & 0.091 &0.087\\
                ours & 0.086 & \textbf{0.086} &\textbf{0.086}\\ 
            \bottomrule
        \end{tabular}
    \end{threeparttable}
    \label{TableOneGap}
\end{table}

\subsection{Illmination \& Elastomer}\label{two gaps}

In this subsection, we study the influence of two domain variables, i.e., illumination condition and elastomer modulus on four domain adaptation groups listed in Table \ref{Domain_Gap_Define}. As shown in Table \ref{TableTwoGaps}, the average errors using the source-only method remain high, around $0.285$ N on average across all four groups. Upon utilizing our model, the average errors improve significantly in all groups, with notable improvements observed in the groups $wmb_1i_1 \rightarrow wmb_0i_0$ and $wmb_0i_0 \rightarrow wmb_1i_1$, where the errors decrease from 0.226 N to 0.095 N and from 0.3 N to 0.138 N respectively. The total average error of $0.145$ N outperforms other models. 

It is noteworthy that when $wmb_1i_1$ is used as the source domain, the force prediction errors in the source-only method are smaller compared to the other two groups. This is because the sensor with elastomer $b_1$ is harder than the counterpart with elastomer $b_0$, resulting in larger contact forces with the same contact depth. Hence, more feature changes induced by larger normal forces, such as elastomer's deformation and light intensity, contains in the source domain, leading to better adaptation performance. Additionally, when $wmb_2i_2$ serves as the target domain, the transfer performance is the poorest, shown in Table \ref{TableTwoGaps} and Fig.~\ref{force_errors}, with the $R^2$ values for the normal force in $wmb_1i_1 \rightarrow wmb_2i_2$ improving from 0.47 to 0.67, compared to the improvement from 0.46 to 0.93 in $wmb_1i_1 \rightarrow wmb_0i_0$. This discrepancy can be attributed to the data size, as $wmb_2i_2$ has ten times fewer images than the other three groups, as mentioned in Section \ref{dataCollect}. This indicates that the size of both target and source data also influences domain adaptation performance.

\begin{table}[htbp]
    \centering
    \caption{Average Force Prediction Error with \textbf{Two} variables\\ (Elastomer \& Illumination, Unit $N$)}
    \centering
    \begin{threeparttable}
        \begin{tabular}{c|cccc|c}
            \toprule
            Methods & \makecell{$wmb_1i_1$  \\ $\rightarrow$ \\ $wmb_2i_2$}& \makecell{$wmb_1i_1$ \\ $\rightarrow$ \\ $wmb_0i_0$}&\makecell{$wmb_0i_0$ \\ $\rightarrow$ \\ $wmb_1i_1$}& \makecell{$wmb_0i_0$ \\ $\rightarrow$ \\ $wmb_2i_2$}& Avg \\
            \midrule
           source-only & 0.230 & 0.226 & 0.300 & 0.384 & 0.285\\
            \midrule
            DANN & 0.180 & 0.120 &\textbf{0.132}&\textbf{0.162}&0.149\\
            GRAM & 0.242 & 0.102 & 0.143 & 0.186 & 0.168\\
            ours & \textbf{0.167} & \textbf{0.095} & 0.138 & 0.180 & \textbf{0.145}\\
            \bottomrule
        \end{tabular}
    \end{threeparttable}
    \label{TableTwoGaps}
\end{table}

On the other hand, although the $R^2$ values for shear forces still remain relatively low, the improvement over the baseline source-only method is significant. In the transition from $wmb_1i_1$ to $wmb_0i_0$, the shear force errors decrease from $0.168$ N (11.2\%) to 0.095 N (6.3\%) in $F_x$ and from 0.195 N (13\%) to 0.093N (6.2\%) in $F_y$. Similarly, in the transition from $wmb_1i_1$ to $wmb_2i_2$, the shear force errors decrease from 0.148 N (9.8\%) to $0.129$ N (8.6\%) in $F_x$ and from 0.180 N (12.0\%) to 0.136 N (9.0\%) in $F_y$. The improvement in normal force is also notable, decreasing from 0.315 N (10.5\%) to 0.097 N (3.2\%) in $wmb_1i_1 \rightarrow wmb_0i_0$, and from 0.361 N (12.0\%) to 0.236 N (7.9\%) in $wmb_1i_1 \rightarrow wmb_2i_2$.

\subsection{Marker Presence \& Illmination \& Elastomer}\label{three gaps}

We finally combine all three domain variables and study three adaptation groups listed in Table \ref{Domain_Gap_Define}. As shown in Table \ref{Table4}, it is evident that with the increase of domain variables, the average force prediction error rises, from 0.236 N (one variable, in Table \ref{TableOneGap}) to 0.285 N (two variables, in Table \ref{TableTwoGaps}) to 0.332 N (three variables) when using the source-only method. This trend implies the increased complexity of domain adaptation performance with more domain gaps. Despite this, our method also outperforms the other two methods and reduces the average error from 0.332 N to 0.153 N. Fig.~\ref{force_errors} demonstrates the transfer performances in the three axis. We observe that, even when combined with three variables, the $R^2$ value in normal force can be improved from -0.39 to 0.89 in $wmb_1i_1 \rightarrow mb_0i_0$. Although the $R^2$ value in $mb_0i_0 \rightarrow wmb_2i_2$ only shows changes from -0.59 to 0.47, the force error decreases from 0.637 N (21.2\%) to $0.299$ N (9.9\%). For the shear force in $wmb_1i_1 \rightarrow mb_0i_0$, the force error get notably improved, with $E$ decreasing from 0.296 N (19.7\%) to 0.098 N (6.5\%) in $x$-axis, from 0.362 N (24.1\%) to 0.103 N (6.9\%) in $y$-axis, and from 0.446 N (14.9\%) to 0.114 N (3.8\%) in $z$-axis. 

\begin{table}[htbp]
    \centering
    \caption{Average Force Prediction Error with \textbf{Three} variables\\ (Maker \& Elastomer \& Illumination, Unit $N$)}
    \centering
    \begin{threeparttable}
        \begin{tabular}{c|ccc|c}
            \toprule
            Methods & \makecell{$mb_0i_0$  \\ $\rightarrow$ \\ $wmb_1i_1$}& \makecell{$mb_0i_0$ \\ $\rightarrow$ \\ $wmb_2i_2$}&\makecell{$wmb_1i_1$ \\ $\rightarrow$ \\ $mb_0i_0$} & Avg \\
            \midrule
            source-only & 0.323 & 0.304 & 0.368 & 0.332\\
            \midrule
            DANN & \textbf{0.143} & 0.238 & 0.185 & 0.189\\
            GRAM & 0.159 & 0.198 & 0.129 & 0.162\\
            ours & 0.160 & \textbf{0.195} & \textbf{0.105} & \textbf{0.153}\\
            \bottomrule
        \end{tabular}
    \end{threeparttable}
    \label{Table4}
\end{table}

\subsection{Discussion}

\begin{figure*}[htbp]
\centering
\includegraphics[width=0.9\textwidth]{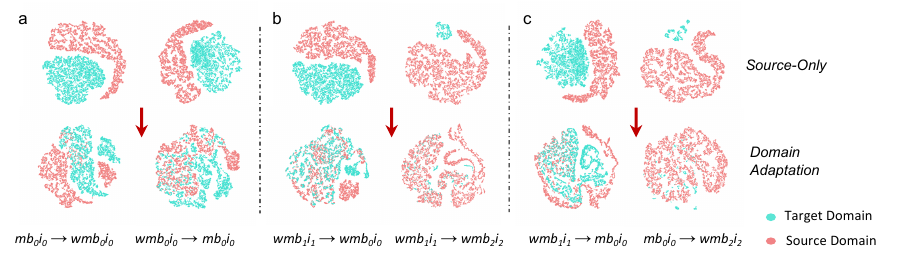}
\caption{Feature spaces visualized using tSNE from the source-only method and our domain adaptation method in (a) one domain variable, (b) two domain variables, (c) and three domain variables.}
\label{figtSNE}
\end{figure*}

The experimental results in Sections \ref{one gap}, \ref{two gaps}, and \ref{three gaps} have validated the feasibility of domain adaptation regression methods in the force calibration of optical tactile sensors. Particularly noteworthy is our method's highest error percentage improvement in group $wmb_0i_0 \rightarrow mb_0i_0$, with improvements of 13.2\% and 17.2\% in shear forces, and 11.1\% in normal force. Additionally, in the $wmb_0i_0 \rightarrow mb_0i_0$ transition, the prediction error can be reduced to lowest among nine groups as 0.102N (3.4\%) in normal force, and 0.095N (6.3\%) and 0.062N (4.1\%) in shear forces. Fig.~\ref{figtSNE} further illustrates that the feature representations of source and target domains are successfully aligned in feature space after domain adaptation, compared with the distinctly separated feature space observed when using the source-only method. As such, considering the real-world deployment of force prediction in robotic manipulation, the prediction accuracy achieved by domain adaptation methods, especially for normal forces, is accurate enough for force feedback control by detecting the threshold of contact forces.

However, our method still exhibits inferior performance in shear force calibration compared with the supervised method. This phenomenon could be attributed to the use of the sphere indenter with low friction coefficient, resulting in slip occurring over a very small moving distance ($<$0.5 mm) thus cannot apply large shear forces on the elastomer. This determines the collected tactile images containing negligible shear features, as they are applied with small lateral forces. Furthermore, the method primarily focuses on accepting static contact images with reference images, which inherently contain poor information about dynamic shear forces. Promising methods to overcome these drawbacks may involve introducing sequential tactile images to provide temporal information about shear displacements.

\section{CONCLUSION}
\label{sec:conclusion}
In this study, we propose a deep domain adaptation regression method for force calibration of optical tactile sensors. This method is promising to reduce the time consumed in the data collection process and eliminate the use of labeled tactile images collected from expensive force/torch sensors. We also study domain gaps of marker presence, illumination conditions, and elastomer modulus on domain transfer performance. We believe that this work not only provides a method for force calibration of GelSight sensors but also holds promise for enhancing the performance of various optical tactile sensors and tactile sensors based on different sensing principles.

\bibliographystyle{IEEEtran}
\bibliography{manuscript}

\begin{thebibliography}{10}
\providecommand{\url}[1]{#1}
\csname url@samestyle\endcsname
\providecommand{\newblock}{\relax}
\providecommand{\bibinfo}[2]{#2}
\providecommand{\BIBentrySTDinterwordspacing}{\spaceskip=0pt\relax}
\providecommand{\BIBentryALTinterwordstretchfactor}{4}
\providecommand{\BIBentryALTinterwordspacing}{\spaceskip=\fontdimen2\font plus
\BIBentryALTinterwordstretchfactor\fontdimen3\font minus \fontdimen4\font\relax}
\providecommand{\BIBforeignlanguage}[2]{{%
\expandafter\ifx\csname l@#1\endcsname\relax
\typeout{** WARNING: IEEEtran.bst: No hyphenation pattern has been}%
\typeout{** loaded for the language `#1'. Using the pattern for}%
\typeout{** the default language instead.}%
\else
\language=\csname l@#1\endcsname
\fi
#2}}
\providecommand{\BIBdecl}{\relax}
\BIBdecl

\bibitem{GelSight1}
W.~Yuan, S.~Dong, and E.~H. Adelson, ``Gel{S}ight: High-resolution robot tactile sensors for estimating geometry and force,'' \emph{Sensors}, vol.~17, no.~12, p. 2762, 2017.

\bibitem{GelSight0}
R.~Li, R.~Platt, W.~Yuan, A.~Ten~Pas, N.~Roscup, M.~A. Srinivasan, and E.~Adelson, ``Localization and manipulation of small parts using {GelS}ight tactile sensing,'' in \emph{IROS}, 2014, pp. 3988--3993.

\bibitem{geltip}
D.~F. Gomes, Z.~Lin, and S.~Luo, ``Gel{T}ip: A finger-shaped optical tactile sensor for robotic manipulation,'' in \emph{IROS}, 2020, pp. 9903--9909.

\bibitem{digit}
M.~Lambeta, P.-W. Chou, G.~Kammerer \emph{et~al.}, ``Digit: A novel design for a low-cost compact high-resolution tactile sensor with application to in-hand manipulation,'' \emph{RA-L}, vol.~5, no.~3, pp. 3838--3845, 2020.

\bibitem{rc1}
Y.~Yan, Z.~Hu, Z.~Yang, W.~Yuan, C.~Song, J.~Pan, and Y.~Shen, ``Soft magnetic skin for super-resolution tactile sensing with force self-decoupling,'' \emph{Science Robotics}, vol.~6, no.~51, p. eabc8801, 2021.

\bibitem{rc2}
Z.~Lu and H.~Yu, ``Gtac-hand: A robotic hand with integrated tactile sensing and extrinsic contact sensing capabilities,'' \emph{IEEE/ASME Transactions on Mechatronics}, vol.~28, no.~5, pp. 2919--2929, 2023.

\bibitem{insight}
H.~Sun, K.~J. Kuchenbecker, and G.~Martius, ``A soft thumb-sized vision-based sensor with accurate all-round force perception,'' \emph{Nature Machine Intelligence}, vol.~4, no.~2, pp. 135--145, 2022.

\bibitem{trans_tatile}
C.~Sferrazza and R.~D’Andrea, ``Transfer learning for vision-based tactile sensing,'' in \emph{IROS}, 2019, pp. 7961--7967.

\bibitem{slip-detect}
W.~Yuan, E.~Adelson \emph{et~al.}, ``Measurement of shear and slip with a {GelSight} tactile sensor,'' in \emph{ICRA}, 2015, pp. 304--311.

\bibitem{texturerecog}
G.~Cao, J.~Jiang, D.~Bollegala, and S.~Luo, ``Learn from incomplete tactile data: Tactile representation learning with masked autoencoders,'' in \emph{IROS}, 2023, pp. 10\,800--10\,805.

\bibitem{localization}
H.~Qi, B.~Yi, S.~Suresh, M.~Lambeta, Y.~Ma, R.~Calandra, and J.~Malik, ``General in-hand object rotation with vision and touch,'' in \emph{Conference on Robot Learning}, 2023, pp. 2549--2564.

\bibitem{uvtac}
W.~Kim, J.~Kim \emph{et~al.}, ``Uvtac: Switchable uv marker-based tactile sensing finger for effective force estimation and object localization,'' \emph{RA-L}, vol.~7, no.~3, pp. 6036--6043, 2022.

\bibitem{gelslim3}
I.~H. Taylor, S.~Dong, and A.~Rodriguez, ``Gel{S}lim 3.0: High-resolution measurement of shape, force and slip in a compact tactile-sensing finger,'' in \emph{ICRA}, 2022, pp. 10\,781--10\,787.

\bibitem{gelwedge}
S.~Wang, Y.~She, B.~Romero, and E.~Adelson, ``{GelSight} wedge: Measuring high-resolution 3d contact geometry with a compact robot finger,'' in \emph{ICRA}, 2021, pp. 6468--6475.

\bibitem{gelsvelte}
J.~Zhao and E.~H. Adelson, ``{GelSight Svelte}: A human finger-shaped single-camera tactile robot finger with large sensing coverage and proprioceptive sensing,'' in \emph{IROS}, 2023, pp. 8979--8984.

\bibitem{hyperelastic}
S.~K. Melly, L.~Liu, Y.~Liu, and J.~Leng, ``A review on material models for isotropic hyperelasticity,'' \emph{International Journal of Mechanical System Dynamics}, vol.~1, no.~1, pp. 71--88, 2021.

\bibitem{DA1}
F.~Zhuang, Z.~Qi, K.~Duan, D.~Xi, Y.~Zhu, H.~Zhu, H.~Xiong, and Q.~He, ``A comprehensive survey on transfer learning,'' \emph{Proceedings of the IEEE}, vol. 109, no.~1, pp. 43--76, 2021.

\bibitem{DANN}
Y.~Ganin and V.~Lempitsky, ``Unsupervised domain adaptation by backpropagation,'' in \emph{ICML}, 2015, pp. 1180--1189.

\bibitem{Coral}
B.~Sun and K.~Saenko, ``Deep coral: Correlation alignment for deep domain adaptation,'' in \emph{ECCV Workshop}, 2016, pp. 443--450.

\bibitem{KL}
M.~Sugiyama, S.~Nakajima \emph{et~al.}, ``Direct importance estimation with model selection and its application to covariate shift adaptation,'' \emph{Advances in neural information processing systems}, vol.~20, 2007.

\bibitem{MMD}
J.~Huang, A.~Gretton, K.~Borgwardt, B.~Sch{\"o}lkopf, and A.~Smola, ``Correcting sample selection bias by unlabeled data,'' \emph{Advances in neural information processing systems}, vol.~19, 2006.

\bibitem{DSAN}
Y.~Zhu, F.~Zhuang, J.~Wang, G.~Ke, J.~Chen, J.~Bian, H.~Xiong, and Q.~He, ``Deep subdomain adaptation network for image classification,'' \emph{IEEE transactions on neural networks and learning systems}, vol.~32, no.~4, pp. 1713--1722, 2020.

\bibitem{seg}
Y.~Liu, W.~Zhang, and J.~Wang, ``Source-free domain adaptation for semantic segmentation,'' in \emph{Proceedings of the IEEE/CVF Conference on Computer Vision and Pattern Recognition}, 2021, pp. 1215--1224.

\bibitem{RSD}
X.~Chen, S.~Wang \emph{et~al.}, ``Representation subspace distance for domain adaptation regression.'' in \emph{ICML}, 2021, pp. 1749--1759.

\bibitem{GRAM}
I.~Nejjar, Q.~Wang, and O.~Fink, ``Dare-gram: Unsupervised domain adaptation regression by aligning inverse gram matrices,'' in \emph{Proceedings of the IEEE/CVF Conference on Computer Vision and Pattern Recognition}, 2023, pp. 11\,744--11\,754.

\bibitem{clfhelpreg}
S.~L. Pintea, Y.~Lin, J.~Dijkstra, and J.~C. van Gemert, ``A step towards understanding why classification helps regression,'' in \emph{ICCV}, 2023, pp. 19\,972--19\,981.

\bibitem{TELEA}
A.~Telea, ``An image inpainting technique based on the fast marching method,'' \emph{Journal of Graphics Tools}, vol.~9, no.~1, pp. 23--34, 2004.

\end{thebibliography}
\end{document}